\documentclass[runningheads,a4paper]{llncs}

\usepackage{amssymb}
\usepackage{graphicx}

\usepackage{url}
\urldef{\mailsa}\path|{alfred.hofmann, ursula.barth, ingrid.haas, frank.holzwarth,|
\urldef{\mailsb}\path|anna.kramer, leonie.kunz, christine.reiss, nicole.sator,|
\urldef{\mailsc}\path|erika.siebert-cole, peter.strasser, lncs}@springer.com|

\usepackage{epsfig}
\usepackage{amsmath}
\usepackage{amssymb}
\usepackage{booktabs}
\def\baselinestretch{0.965}

\usepackage{caption}
\usepackage{subcaption}
\usepackage{breakcites}

\def\onedot{. }
\def\eg{\emph{e.g}\onedot} 
\def\ie{\emph{i.e}\onedot}

\def\etal{\emph{et al}\onedot}

\begin{document}

\mainmatter  

\title{Holistically-Nested Structure-Aware Graph Neural Network for Road Extraction}

\titlerunning{Holistically-Nested Structure-Aware Graph Neural Network for Road Extraction}

\authorrunning{Tinghuai Wang et al.}

\author{%
	Tinghuai Wang, Guangming Wang, Kuan Eeik Tan
}%

\author{%
	Tinghuai Wang \and 
	Guangming Wang\and 
	Kuan Eeik Tan
}%
\institute{
	Huawei Helsinki Research Center, Finland}


%
%
%



%
%

\maketitle

\begin{abstract}

Convolutional neural networks (CNN) have made significant advances in detecting roads from satellite images. However,
existing CNN approaches are generally repurposed semantic segmentation architectures and suffer from the poor delineation of long and curved regions. Lack of overall road topology and structure information further deteriorates their performance on challenging remote sensing images. This paper presents a novel multi-task graph neural network (GNN) which simultaneously detects both road regions and road borders; the inter-play between these two tasks unlocks superior performance from two perspectives: (1)  the hierarchically detected road borders enable the network to capture and encode holistic road structure to enhance road connectivity (2) identifying the intrinsic correlation of semantic landcover regions mitigates the difficulty in recognizing roads cluttered by regions with similar appearance. Experiments on challenging dataset demonstrate that the proposed architecture can improve the road border delineation and road extraction accuracy compared with the existing methods.

\end{abstract}

\section{Introduction}
\label{sec:intro}
Accurate road extraction from high-resolution satellite images has become critical in various geospatial applications such as cartography, map updating, urban planning, and navigation. The high-resolution images have posed new challenges for road extraction methods by presenting more details, such as multiscale roads and complex background. 

Recent years has witnessed the tremendous success of applying deep convolutional neural networks (CNNs) in tackling this challenge, by formulating road extraction as a binary segmentation problem. To this end, existing CNN based methods largely adopt semantic image segmentation neural architectures which are repurposed for road extraction task. None-the-less, these architectures normally suffer from two major drawbacks: (1) semantic segmentation architectures are not optimized to recognize thin, long and curved shapes which are frequently mis-segmented due to the nature of spatial downsampling operations of convolutional neural networks; this lack of contour modeling limits their ability to precisely localize road borders which however are the key information for generating map data (2) these architectures focus on per-pixel prediction gathering local context however road topology and holistic road structure information are not utilized; road regions are sparsely distributed over the broad spatial domain cluttered by trees, shadows and buildings etc – in the representation of deep features, some road fragments might be easier to be detected than other road fragments, and thus road topology and holistic structure information should be incorporated to enhance road connectivity in designing deep neural networks.  

In this paper, we present a novel multi-task graph neural network (GNN) which simultaneously detects both road regions and road borders; the inter-play between the two tasks unlocks superior performance from two perspectives: (1)  the hierarchically detected road borders enable the network to capture and encode holistic road structure to enhance road connectivity (2) identifying the intrinsic correlation of landcover regions mitigates the difficulty in recognizing roads cluttered by regions with similar appearance, such as roads around parking area. 

To this end, this paper makes the following novel contributions: 
\begin{itemize}
    \item We propose to hierarchically detect road borders in an end-to-end manner in order to emphasizes morphological feature learning and border localization
    \item We propose a novel architecture to reason the interplay between road regions and borders, which consists of a set of nested parallel graph convolutional networks with one branch harnessing morphological feature which inferences the connectivity of detected road regions, and the other branch  capturing the correlation among border features represented in landcover feature space which in turn discriminates subtle borders which can only be inferred based on contextual information of landcover regions.
\end{itemize}

\section{Related Work}
\label{sec:relatedwork}

A variety of methods for road extraction have been proposed in recent years. Road extraction is generally considered
as a binary segmentation or pixel-level classification problem, and machine learning algorithms have been applied. For instance, support vector machines (SVMs) have been utilized to detect roads by classifying hand-crafted features \eg  shape index features \cite{song2004road} or salient features \cite{DasMV11}; graph optimization approach has also been adopted \eg  \cite{alshehhi2017hierarchical} based on the hierarchical graph-based image segmentation work \cite{felzenszwalb2004efficient}.

The past decade has witnessed the remarkable success of deep learning in various challenging computer vision and remote sensing \cite{WangW14,WangW16,ZhuWAK19,zhu2019cross,Wang20,WangWTT20,tinghuai2020watermark,jia21hsi,jia21road,jiang2022convolutional,tinghuai2022method,jiang2022implementation,jiang2022smartphone} tasks. One of the first deep learning based road extraction methods was developed by Mnih and Hinton \cite{mnih2010learning}, which proposed to adopt restricted Boltzmann machines (RBMs) to detect road from high resolution aerial images. Saito \etal \cite{saito2016multiple} applied Convolutional Neural Network (CNN) to extract both buildings and roads from satellite images without pre-processing step required by \cite{mnih2010learning}.
A U-shaped fully convolutional network (FCN) \cite{long2015fully} architecture was proposed by Ramesh \etal \cite{kestur2018ufcn}. This architecture comprised a stack of convolutions followed by corresponding stack of mirrored deconvolutions with the usage of skip connections in between for preserving the local information. Zhang \etal \cite{zhang2018road} combined the strengths of U-Net \cite{ronneberger2015u} and residual learning \cite{he2016deep} in training deep neural networks for road extraction. Methods in tackling the small receptive field issue have also been proposed. For instance, Zhou \etal \cite{zhou2018d} built upon the dilated convolution \cite{chen2017rethinking} layers to adjust receptive fields of feature points without decreasing the resolution of feature maps. A comprehensive review of road extraction methods can be found in \cite{jia2021review}. 

Multi-task learning has been adopted for improving the accuracy of road extractions. Liu \etal \cite{liu2018roadnet} developed multi-task learning network to extract road surfaces, centerlines, and edges simultaneously using FCN networks. Lu \etal \cite{lu2019multi} utilized U-Net with two prediction outputs to perform both road detection and centerline extraction tasks. Yang \etal \cite{yang2019road} proposed a similar U-Net type network to perform simultaneous prediction of road detection and centerline extraction tasks. Shao \etal \cite{shao2021mrenet} proposed a two-task network which contains two joint U-Net type networks for both road detection and centerline extraction. 
The focus of these multi-task networks is to produce multiple predictions adapting U-Net type networks rather than exploring the intrinsic interplay among multiple predictions as in our work. The fact that existing methods utilize CNN architectures further limit their capability to embed topology information and reason over longer range connectivities. 


The connectivity in road extraction has attracted considerable research efforts. Chen \etal \cite{chen2018road} proposed a two-stage method which applied connection analysis on the discrete line features with directional consistency to extract potential road objects and evaluate potential road objects using shape features to refine the road extraction results. Gao \etal \cite{gao2019road}, another two-stage method, used semantic segmentation to obtain pixel-level road segmentation results, and then used tensor voting to connect broken roads. Oner \etal \cite{oner2020promoting} proposed a differentiable loss function which enforced connectivity on the output of binary segmentation N-Net for the purpose of road network delineation. 


Graphical models and graph neural networks (GNNs) have emerged as powerful tools in the field of computer vision 
\cite{wang2010multi,wang2017submodular,qi20173d,tinghuai2018method1,
wang2019zero,zhu2019portrait,wang2019graph,tinghuai2020semantic,yang2021learning}, offering a versatile framework for representing and processing complex visual data. 
These approaches leverage the inherent structure and relationships within images 
\cite{wang2015robust,tinghuai2016method,tinghuai2017method,wang2020spectral,xing2021learning,han2022vision} and videos 
\cite{wang2010video,wang2014wide,tinghuai2018method2,chen2020fine}, allowing for more nuanced and context-aware analysis. By representing visual elements as nodes and their interactions as edges, graphical models and GNNs can capture spatial, temporal, and semantic dependencies that are crucial for tasks such as visual information retrieval \cite{hu2013markov}, stylization \cite{WangCSCG10,wang2011stylized,wang2013learnable}, object detection \cite{tinghuai2016apparatus,zhao2021graphfpn}, scene understanding \cite{deng2021generative}, and image or video segmentation 
\cite{wang2015weakly,wang2011probabilistic,lu2020video,wang2021end}. In recent years, these techniques have also found significant applications in remote sensing and geospatial data analysis.

\begin{figure*}[!t]
	\centering
	\includegraphics[width=\linewidth]{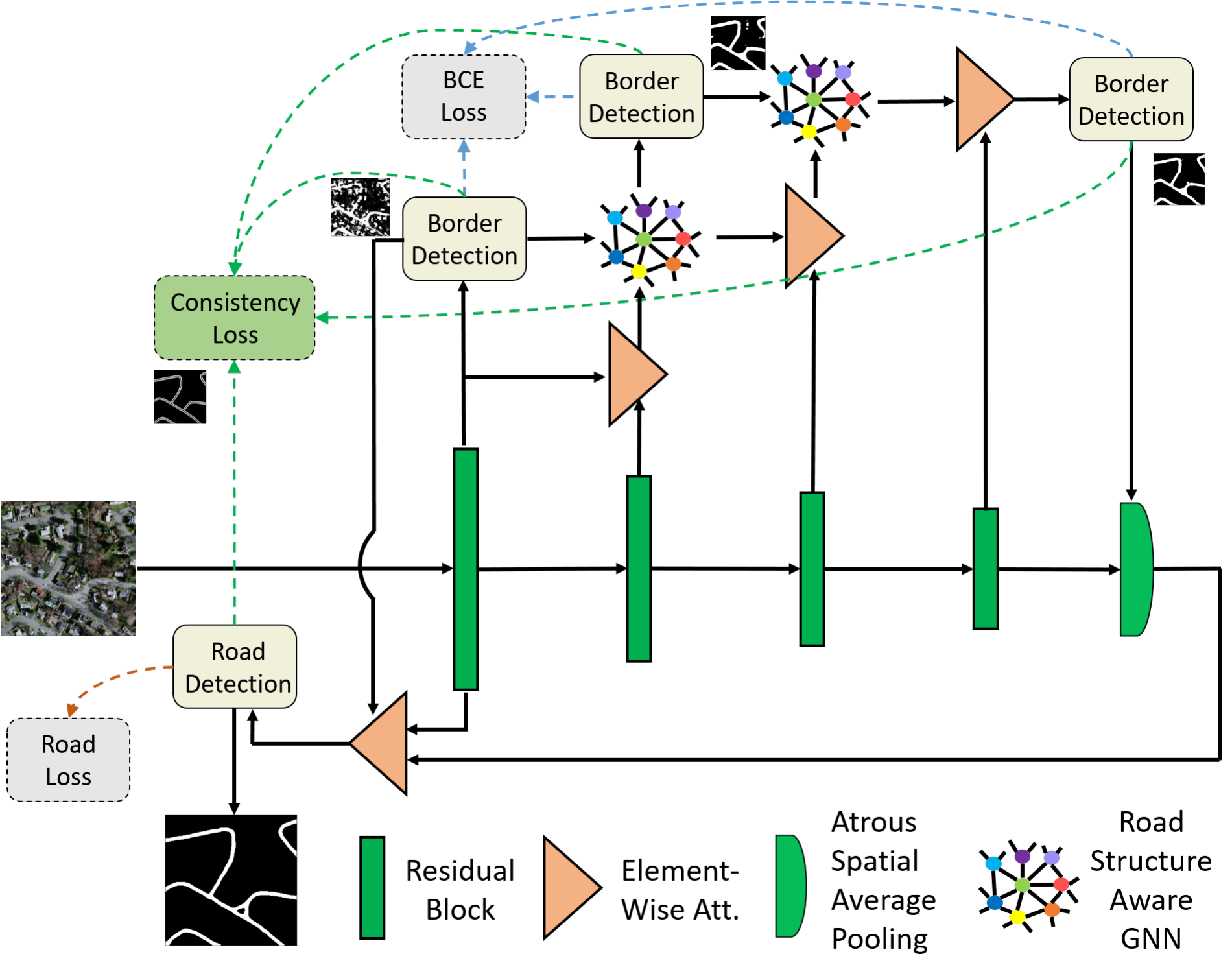}
	\caption{An overview of our proposed architecture.}
	\label{fig:overview}
\end{figure*}

\begin{figure*}[!t]
	\centering
	\includegraphics[width=\linewidth]{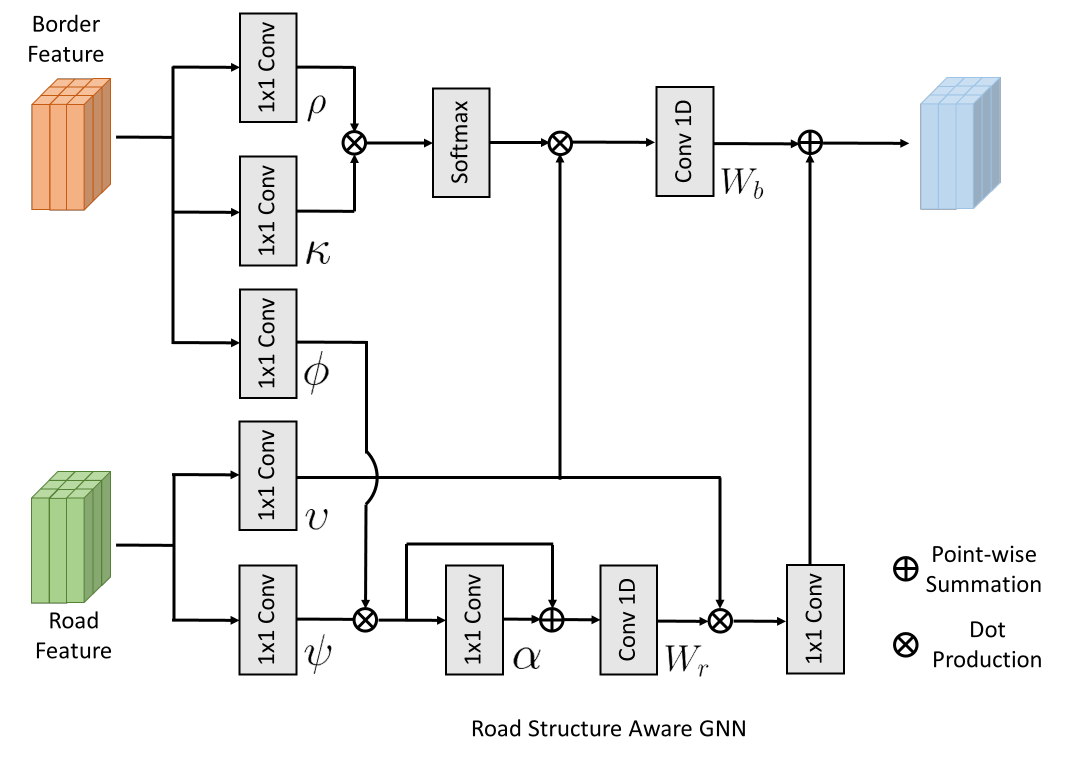}
	\caption{Illustration of road structure aware GNN.}
	\label{fig:gnn}
\end{figure*}

\begin{figure*}[!t]
	\centering
	\includegraphics[width=\linewidth]{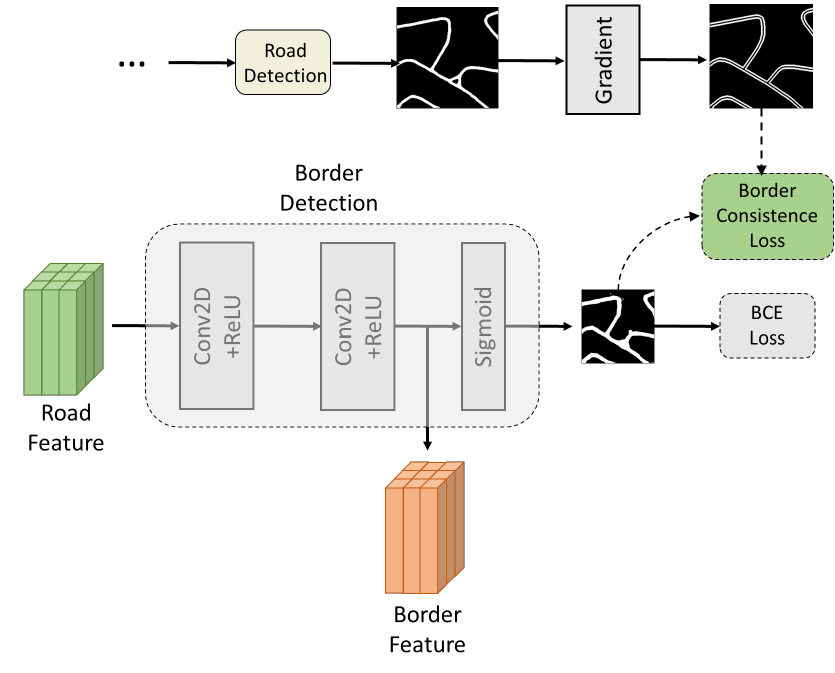}
	\caption{Illustration of border detection module and border consistence loss.}
	\label{fig:loss}
\end{figure*}

\section{Method}

In this section, we present our holistically-nested structure-aware graph neural network architecture for road extraction. As illustrated in Fig. \ref{fig:overview}, our network consists of two streams of networks, \ie encoder network and road structure inference network.
The encoder comprises of four residual blocks according to the size of feature maps, whilst the rest of the blocks constitute the road structure inference network. Encoder extracts appearance and contextual information at various hierarchies, with decreasing spatial details and increasing semantic information from encoding block-1, -2, -3 to block-4. At the output of each encoding block, the encoder interfaces with the proposed road structure inference network by providing feature map as input. Road structure inference network mainly comprises of road border detection, road structure aware GNNs and element-wise attention modules. 

\subsection{Road border detection}

Road border detection module takes feature map, \ie road feature, as input at a certain hierarchy level and estimates road borders. As depicted in \ref{fig:loss}, border detection module consists of two convolutional layers and a softmax layer. Border detection is achieved through deep supervision. Specifically we generate ground-truth data by extracting borders from road masks in the training data, then we use supervised binary cross entropy loss on output borders to supervise the training of road border detection. The feature maps immediately before sigmoid layer are considered as border features which encode key information of road borders.

\subsection{Road structure aware GNN}

Road structure aware GNN is the core module of our network which inferences road regions by seamlessly fusing holistic road structure and  appearance information via graph inference.  Road structure aware GNN module comprises two streams of GNN architectures, as illustrated in Fig. \ref{fig:gnn}. This module takes border feature and road feature as input. Road feature is the fusion of feature maps from adjacent encoding blocks via element-wise attention module described in later section, which contains the semantic representation of regions. 

As shown in Fig. \ref{fig:gnn}, the upper stream is utilizing a self-attention mechanism where a query and a set of key-value pairs are generated. Specifically, border feature, denoted as $\mathbf{X_b}$, is transformed into a query feature $\rho(\mathbf{X_b})$ and a key feature $\kappa(\mathbf{X_b})$, and road feature, denoted as $\mathbf{X_r}$, is transformed into a value feature $\upsilon(\mathbf{X_r})$, through learnable linear transformations  $\rho$, $\kappa$ and $\upsilon$ respectively. Following the scaled dot-product self-attention \cite{waswani2017attention}, we define a co-attention embedding operation as
\begin{align}
\text{softmax} (\frac{\rho(\mathbf{X_b}) \kappa(\mathbf{X_b})^T}{\sqrt{d}}) \upsilon(\mathbf{X_r}),
\label{eq:1}
\end{align} 
where $d$ is dimension of query and key. The softmax dot-product operation scales to have unit norm which is equivalent to cosine similarity used as the adjacency matrix. The significance of this operation is that for each border feature in the query, the weighted sum of value features, \ie semantic road features, is taken as its soft nearest neighbor in border space. Therefore, the output of this operation can be seen as feature in the semantic road feature space but the information of query, \ie border feature, is embedded. This promotes the semantic road feature to flow with an awareness of the intrinsic structure of road borders.  

The lower stream further takes two features as input, namely border feature $\phi(\mathbf{X_b}) \in \mathbb{R}^{N \times D_1}$ and semantic road feature $\psi(\mathbf{X_r}) \in \mathbb{R}^{N \times D_2}$ through learnable linear transformations $\phi$ and $\psi$ respectively. Multiplying these two features 
\begin{align}
\mathbf{X_f} = \phi(\mathbf{X_b})^T \psi(\mathbf{X_r})
\label{eq:2}
\end{align} 
results in a new feature $\mathbf{X_f}\in \mathbb{R}^{D_1 \times D_2}$, which consists of $D_1$ nodes, each of dimension $D_2$. This operation projects each border feature dimension in semantic road feature space. 

After projection, a graph $\mathcal{G}$ with adjacency matrix $\mathbf{A}_{\mathcal{G}} \in \mathbb{R}^{D_1 \times D_1}$ in the road feature space, where each node contains the semantic road feature. Denoting the trainable weights of convolutional layer as $\mathbf{W}_r \in \mathbb{R}^{D_2 \times D_2}$, the GNN convolution is defined as 
\begin{align}
\mathbf{X_l} = (\mathbf{I}-\mathbf{A}_{\mathcal{G}} ) \mathbf{X_f} \mathbf{W}_r
\end{align}
Laplacian smoothing  \cite{wang2014graph,WangW16,wang2016semi,wang2016primary,wang2017cross,li2018deeper} is adopted by updating the adjacency matrix to $\mathbf{I}-\mathbf{A}_{\mathcal{G}}$ to propagate the node features over the graph. Rather than estimating adjacency matrix in the upper stream, we randomly initialize and optimize $\mathbf{A}_{\mathcal{G}} $  by gradient descent in an end-to-end fashion. After the graph reasoning, feature $\mathbf{X_l} $ is projected back to the original road feature space by multiplying $\upsilon(\mathbf{X_r})$. After another convolutional layer, the new feature is fused with feature from upper stream by point-wise summation. Intuitively, the lower stream identifies the intrinsic correlation of semantic landcover regions which mitigates the difficulty in recognizing roads cluttered by regions with similar appearance, such as roads around parking area.
\subsection{Element-wise attention}

Element-wise attention module is used in multiple locations to fuse the features between neighboring encoding blocks or features from road structure aware GNN and encoding block. Specifically, an attention map $\alpha \in \mathbb{R}^{H \times W}$ is obtained by concatenating two feature maps which is followed by a $1 \times 1$ convolutional layer and a sigmoid layer for normalization. For a feature $\mathbf{X_e}$ from later encoding block, the element-wise attention is computed as
\begin{align}
\mathbf{\hat{X}_e} = (\mathbf{X_e} 	\odot \alpha  + \mathbf{X_e} )  \mathbf{W}_e.
\end{align}
Intuitively, the attention map carries important lower level border information to weigh the area of feature maps with higher semantic information. 

\begin{figure*}[!t]
	\centering
	\includegraphics[width=\linewidth,height=17cm]{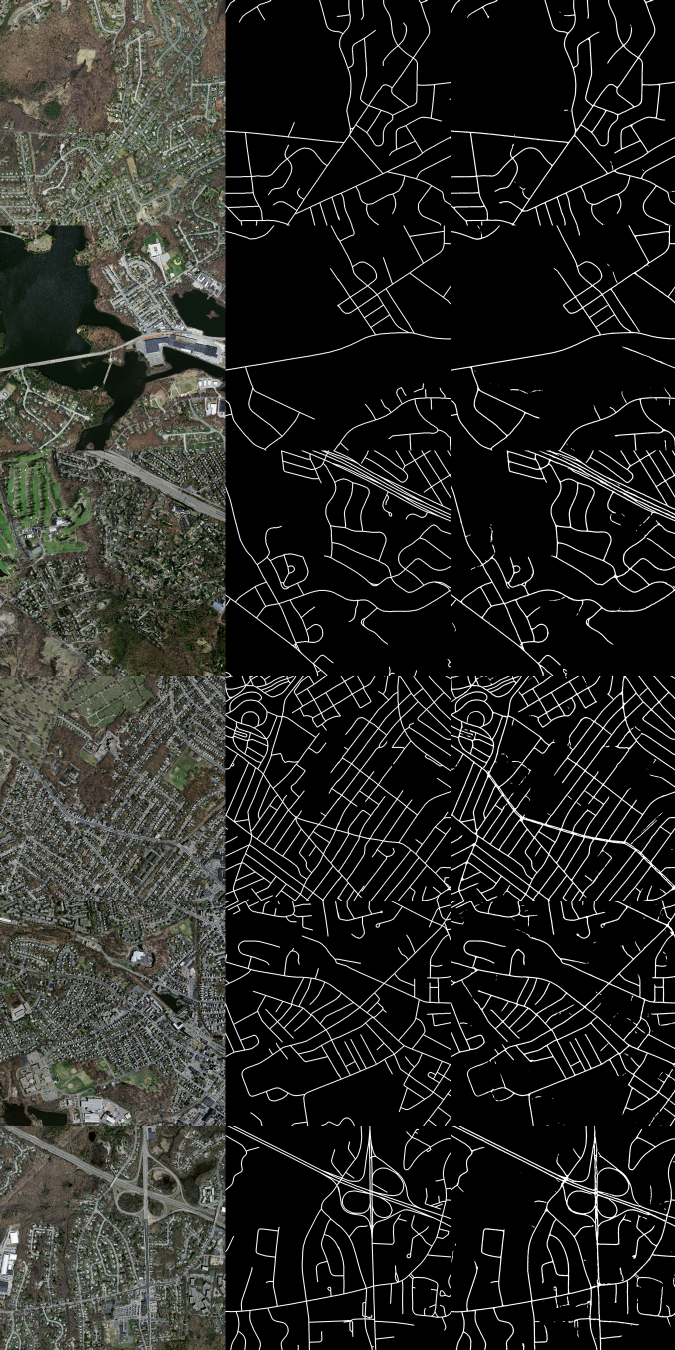}
	\caption{Qualitative results of proposed architecture on Massachusetts road dataset. Left: source image; middle: ground-truth; right: road prediction results.}
	\label{fig:massroads}
\end{figure*}

\subsection{Joint multi-task training}

We jointly supervise road border detection and road region prediction in an end-to-end fashion. We use standard binary cross-entropy (BCE) loss on both predicted border map $b$ and predicted road region $y$,
\begin{align}
\mathcal{L}_r = \mathcal{L}_{BCE} (b, \hat{b}) + \mathcal{L}_{BCE} (y, \hat{y})
\end{align}
where $\hat{b})$ and  $\hat{y})$ denote road border and road region ground-truth labels. The high imbalance between road and non-road, as well as border and non-border pixels is also properly compensated following \cite{xie2015holistically}.

We further define a border consistence loss by measuring the dependency between the predicted border and the border of predicted road region,
 \begin{align}
 \mathcal{L}_c^i = \frac{1}{|\mathcal{N}^+|}  \sum |\frac{1}{\sqrt{2}} ||\bigtriangledown \hat{y}|| - \hat{b}^i|
 \end{align}
where $\mathcal{N}^+$ denotes the union of border pixels in both border maps, superscript $i$ indicates the loss defined at layer $i$ border prediction, $\bigtriangledown$ represents the spatial derivative operation.

\section{Experimental results}
In this section, we conduct extensive experiments on a publicly available aerial imagery dataset for roads in Massachusetts, USA \cite{mnih2013machine}. This dataset provides 1171 images including 1108 training images, 49 test images, and 14 validation images. The spatial resolution is 1.2 m for all images which are $1500 \times 1500$ pixels. This dataset is a challenging aerial image labeling dataset since the images cover an area of 2600 $km^2$ with a wide variety of geographic types including urban, suburban and rural regions. 

Similar with previous work, \eg \cite{xu2018road} , we use random crops of size $256\times 256$ pixels from training set for training the network whilst we directly test on the test images at $1500\times 1500$ pixels without any processing. We train the network with a batch size of 20 for 50 epochs, using Adam optimizer with learning rate of $10^{-3}$. 

\subsection{Evaluation metrics}
We used both IoU and F1-score to evaluate quantitative performance. The IoU metric measures the intersection of the prediction and ground truth regions over their union,
 \begin{align}
\text{IoU} = \frac{\text{TP}}{\text{TP+FP+FN}},  \nonumber
 \end{align}
where $\text{TP}$, $\text{FP}$ and $\text{FN}$ represent the number of true positives, false positives and false negatives respectively. F1-score is also used which is defined as 
 \begin{align}
 \text{F1} = \frac{2\cdot\text{precision}\cdot\text{recall}}{\text{precision}+\text{recall}}\\\nonumber
 \text{recall} = \frac{\text{TP}}{\text{TP+FN}}, ~ \text{precision} = \frac{\text{TP}}{\text{TP+FP}}.\nonumber
 \end{align}
 
In addition to the above region based metrics, we follow a boundary metric \cite{perazzi2016benchmark} to evaluate the quality of our road predictions. This metric computes the F-score along the boundary of the predicted mask, given a small relax in distance. In our experiments, we use thresholds 1, 2, 3, 4, 5 pixels respectively.

\subsection{Results}

We quantitatively and qualitatively evaluate our proposed method and compare with various recent deep learning based methods to demonstrate its effectiveness.

Table \ref{table:mass} presents the quantitative results obtained by different methods on the Massachusetts dataset. Our proposed approach outperforms all the competing state-of-the-art neural architectures. As opposed to the general convolutional neural architectures adopted by the compared methods, our method is the only graph neural network architecture for road extraction. We owe the superior performance of our method to the multiple graph reasoning across the information from multi-task learning. Specifically, we construct a set of nested parallel graph convolutional networks with one stream harnessing morphological feature which inferences the connectivity of detected road regions, and the other stream capturing the correlation among border features represented in semantic feature space which in turn discriminates subtle borders which can only be inferred based on contextual information of landcover regions. 

\begin{figure*}[!t]
	\centering
	\includegraphics[width=\linewidth]{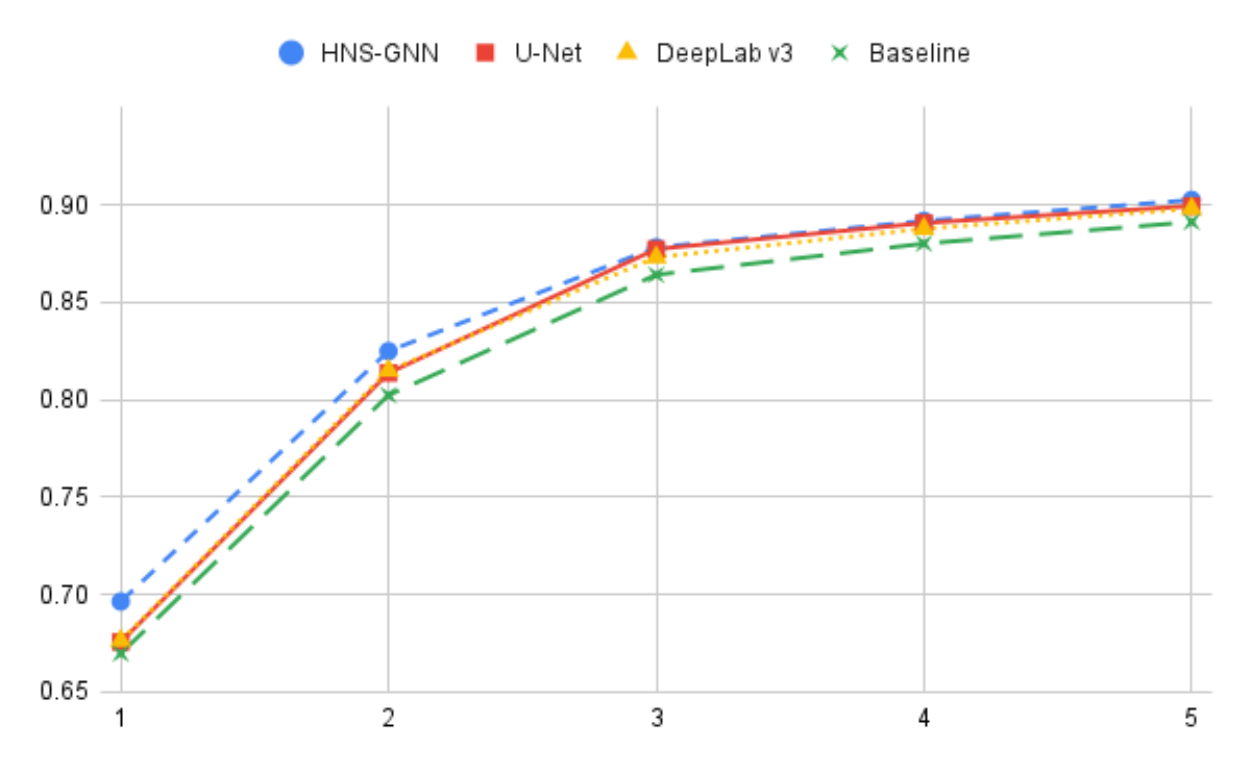}
	\caption{Comparison vs baselines at different thresholds in terms of boundary F-score on Massachusetts road dataset test set.}
	\label{fig:chart}
\end{figure*}

Fig. \ref{fig:massroads} shows visual results of our proposed architecture where superior border accuracy and road connectivity of predictions can be clearly observed. Such excellent border delineation is quantitatively confirmed by comparing with state-of-the-art architectures in terms of boundary F1-score as shown in Fig. \ref{fig:chart}, where our proposed method consistently demonstrates higher boundary accuracy regardless of varying thresholds and larger margin in the case of 1-pixel threshold. 

We also conduct abalation study to investigate the contribution of each proposed modules. The first baseline network, \ie HNS-GNN-BU, removes 
border detection and road structure aware GNN modules and only keeps element-wise attention based feature fusion,
which attains F1-score of 74.95\%. Comparing with our final architecture, \ie HNS-GNN, our core contribution of multitask learning and graph reasoning increases the accuracy by 2.6\%.  The second baseline, \ie HNS-GNN-SG, adds the upper stream GNN to HNS-GNN-BU which increases 0.96\%. The third and fourth baseline, \ie HNS-GNN-B1 and HNS-GNN-B2, corresponding to keeping one or two border detection and GNN modules respectively, give accuracy gains of 1.1\% and 2.58\% respectively.

\begin{table*}[!t]
	\centering
	\begin{small}
	\begin{tabular}{lccccccc}
		\hline
		Metric  (\%)   & U-Net \cite{ronneberger2015u}  &   DeepLab v3 \cite{chen2017rethinking} & D-LinkNet \cite{zhou2018d} & GL-Dense-UNet \cite{xu2018road} & CDG \cite{wang2020improved} & Ours\\
		\hline
         IoU & 60.94 & 60.52  & 60.71 & 60.93 & 61.90 &62.94  \\
        F1-Score  & 75.24  & 75.81 & 75.15  & 75.72 & 76.10 & 76.96\\
		\hline
	\end{tabular}
	\end{small}
	\caption{Comparisons of the proposed and other deep learning based road extraction methods on Massachusetts road dataset test set.}
	\label{table:mass}
\end{table*}

\begin{table*}[!t]
	\centering
	\begin{small}
		\begin{tabular}{lccccccc}
			\hline
			Metric  & HNS-GNN-BU  &   HNS-GNN-SG & HNS-GNN-E1   & HNS-GNN-E2    & Ours\\
			\hline
			F1-Score  (\%)  & 74.95 & 75.67  & 75.78 & 76.89 & 76.96 \\
			\hline
		\end{tabular}
	\end{small}
	\caption{Ablation studies of our proposed architecture on Massachusetts road dataset.}
	\label{table:ablation}
\end{table*}

\section{Conclusions}
In this paper, we proposed a novel architecture for extracting road from satellite images. This multi-task graph neural network is able to 
hierarchically detect road borders which in turn enable the network to capture and encode holistic road structure to enhance road connectivity. 
Graph reasoning further identifies the intrinsic correlation of semantic landcover regions which mitigates the difficulty in recognizing roads cluttered by regions with similar appearance.

\bibliographystyle{splncs03}
\bibliography{egbib}
\end{document}